\documentclass[conference]{IEEEtran}
\ifCLASSINFOpdf
\else
\fi
\usepackage{amsmath}
\usepackage{amsfonts}       

\usepackage{threeparttable}

\usepackage{graphicx}
\usepackage{algorithm}
\usepackage{algorithmicx}
\usepackage{algpseudocode}%
\usepackage{amsmath} 
 
\usepackage{booktabs}
\usepackage{multirow}
\usepackage{multicol}
\usepackage{makecell}

\floatname{algorithm}{Algorithm} 

\usepackage{fancyhdr}

\begin{document}
%
\title{Generative Data Augmentation for Non-IID Problem in Decentralized Clinical Machine Learning}


\author{
	\IEEEauthorblockN{
	    Zirui Wang\IEEEauthorrefmark{1},
	    Shaoming Duan\IEEEauthorrefmark{1},
		Chengyue Wu\IEEEauthorrefmark{1},
		Wenhao Lin\IEEEauthorrefmark{1},
		Xinyu Zha\IEEEauthorrefmark{1},
		Peiyi Han\IEEEauthorrefmark{1}\IEEEauthorrefmark{2},
		Chuanyi Liu\IEEEauthorrefmark{1}\IEEEauthorrefmark{2}\IEEEauthorrefmark{3} 
	}
	 
	\IEEEauthorblockA{\IEEEauthorrefmark{1} School of Computer Science and Technology, Harbin Institute of Technology (Shenzhen), Shenzhen, China }
	\IEEEauthorblockA{\IEEEauthorrefmark{2} Cyberspace Security Research Center, Peng Cheng Laboratory, Shenzhen, China }
	\IEEEauthorblockA{\IEEEauthorrefmark{3}Corresponding authors \\
	    Email:854102781@qq.com, shaomingduan@gmail.com,
	    1190201314@stu.hit.edu.cn, \\
	    190110219@stu.hit.edu.cn,
	    200110710@stu.hit.edu.cn,
	    hanpeiyi@hit.edu.cn,
	    liuchuanyi@hit.edu.cn}

}

\maketitle

\begin{abstract}

Swarm learning (SL) is an emerging promising decentralized machine learning paradigm and has achieved high performance in clinical applications. SL solves the problem of a central structure in federated learning by combining edge computing and blockchain-based peer-to-peer network. While there are promising results in the assumption of the independent and identically distributed (IID) data across participants, SL suffers from performance degradation as the degree of the non-IID data increases. To address this problem, we propose a generative augmentation framework in swarm learning called SL-GAN, which augments the non-IID data by generating the synthetic data from participants. SL-GAN trains generators and discriminators locally, and periodically aggregation via a randomly elected coordinator in SL network. Under the standard assumptions, we theoretically prove the convergence of SL-GAN using stochastic approximations. Experimental results demonstrate that SL-GAN outperforms state-of-art methods on three real world clinical datasets including Tuberculosis, Leukemia, COVID-19.

\end{abstract}


\IEEEpeerreviewmaketitle

\section{Introduction}

Machine learning (ML) models are data hungry. In precision medicine \cite{haendel2018classification}, the performance of ML models that identify patients with life-threatening diseases, such as leukemia, tuberculosis or COVID-19, increases with the size and diversity of the training samples \cite{echle2020clinical}. In practice, to train a robust clinical ML model, patient-related data often needs to be centralized in a central repository \cite{chen2021synthetic, howard2021impact}. However, such data sharing across different institutions or countries, faces privacy and legal obstacles. This problem has been solved by federated learning \cite{mcmahan2017communication}, in which multiple participants jointly train a ML model under a central coordinator. In federated learning, each participant trains a local model using the local data separately, and then shares the learned model gradients to the coordinator for model aggregation. Although, the private data is distributed and not disclosing to others, the central structure in federated learning remains vulnerable to attack \cite{chen2021privacy}.

\begin{figure}[!t] 
\centering 
\includegraphics[width=\linewidth]{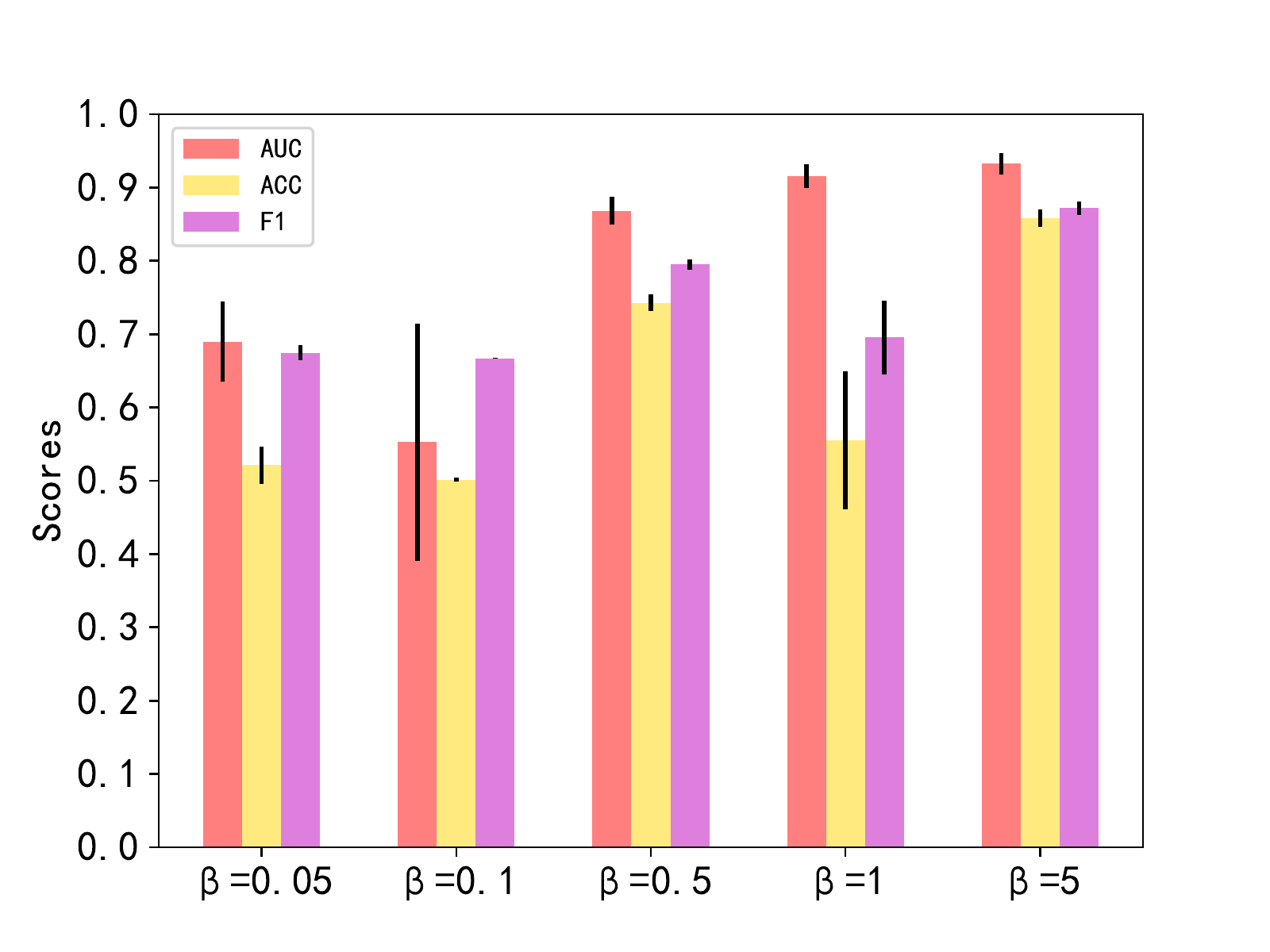} 
\caption{Performance of swarm learning on the Tuberculosis dataset. $\beta$ controls the degree of the non-IID, the lower the $\beta$, the more imbalanced the data distribution is. With the $\beta$ increases, the performance of SL increases.} 
\label{fig:introduction} 
\end{figure}

To tackle the limitations of federated learning, swarm learning (SL) \cite{warnat2021swarm} combines edge computing and blockchain-based peer-to-peer network. In SL, a new participant register via a blockchain smart contract, obtains the global model, and trains the model locally. After a user-defined synchronization interval, local model parameters are exchanged and merged to update the global model by a randomly elected edge node. The chosen edge node replaces the role of central coordinator in federated learning. In the context of clinical data mining, SL provides a fairness environment for multi-parties ML model training and creates a strong incentive to collaborate without data sharing. As SL secures data sovereignty, security, and confidentiality, it has been successfully applied in healthcare fields \cite{warnat2021swarm, saldanha2022swarm, fan2021fairness}. However, the non-independent and identical distributed (non-IID) data heavily limit the performance of SL. As shown in Figure \ref{fig:introduction}, with the degree of non-IID increases ($\beta$ from 5 to 0.05), the AUC scores of SL decreases from 93\% to 69\%.

 Existing methods for solving the non-IID problem in decentralized learning can be roughly divided into two categories: \textit{algorithm-based methods} and \textit{data-based methods}. Algorithm-based methods improve the model robustness by modifying the local loss function \cite{li2020federated, wang2020tackling} to make the local model consistent with the global model, designing a new aggregation scheme \cite{yurochkin2019bayesian, wang2019federated} to improve the model aggregation mechanism, or training personalized models \cite{smith2017federated, li2021ditto} for each participant rather than the same global model. However, as discussed in \cite{li2021federated}, existing algorithm-based methods are not always better than vanilla FedAvg \cite{mcmahan2017communication}. Data-based methods construct a more balanced data distribution among participants or on the server by data sharing or augmentation strategies \cite{zhang2018mixup, rasouli2020fedgan}, which achieve a high performance on non-IID data. Unfortunately, there are still two challenges in  directly applying these methods to SL. On the one hand, these methods need a trusted central coordinator to employ data sharing or data augmentation, which is contrary to the assumption in SL. On the other hand, data augmentation strategies based on generative adversarial network (GAN) remains a challenge is that GAN may not converge on non-IID data. To the best of our knowledge, there are currently no works to solve the non-IID problem in SL.

To address these challenges, we propose a novel generative augmentation framework in SL called SL-GAN, which augments the non-IID data to a balanced data distribution among each participant by using a generative model. In SL-GAN, discriminators and generators are trained locally and aggregated after a user-defined synchronization interval by a randomly elected edge node. Furthermore, under the standard assumption in SL, we theoretically prove the convergence of SL-GAN using stochastic approximations. We evaluate our SL-GAN on three real-world clinical datasets with various data distributions among participants. The experimental results show that SL-GAN can effectively improve the performance of SL on non-IID data and outperforms the state-of-the-art methods.

The main contributions of this paper are as follows:
\begin{enumerate}
\item[(1)] To the best of our knowledge, this is the first study on the non-IID problem in SL. We propose SL-GAN, a novel data augmentation framework in SL, which jointly trains a global generative model to augment the non-IID data without a central coordinator.
\item[(2)] We theoretically prove our SL-GAN converges with non-IID data under the standard assumption in SL.
\item[(3)] We test SL-GAN on three real-world clinical datasets with various data distributions. The experimental results demonstrate that SL-GAN outperforms the state-of-the-art approaches and is robust to varies data distributions.
\end{enumerate}


\section{Related Works}
\label{sec:related}

\subsection{Non-IID in Decentralized Learning}

Since there are currently no works to solve the non-IID problem in swarm learning, we describe the related works in other decentralized learning method, such as federated learning. Existing methods in federated learning to deal with the non-IID problem are divided into two categories \cite{zhu2021federated}: \textit{algorithm-based methods} and \textit{data-based methods}.


\subsubsection{Algorithm-based methods} As suggested in \cite{zhao2018federated}, the weight divergence of local models caused by non-IID data is the root cause of model performance degradation. To alleviate this problem, Fedprox\cite{li2020federated} adds a penalty term in the objective function to make the local model consistent with the global model.
SCAFFOLD\cite{karimireddy2020scaffold} controls the similarity between the local model and the global model by adding a regularization term to the local loss function. To avoid the influence of the local model from nodes with large data volumes, FedNova\cite{wang2020tackling} normalizes the local model before model aggregation. Unlike modifying the objective, personalized federated learning \cite{smith2017federated, li2021ditto} aims to train personalized models in each participant rather than the same global model. However, as shown in \cite{li2021federated}, existing algorithm-based methods are not always better than vanilla FedAvg \cite{mcmahan2017communication}.


\subsubsection{Data-based methods}
To construct a balanced data distribution among each participant, data-based methods perform data augmentation or data sharing strategies under the coordination of a central server. Mixup\cite{zhang2018mixup} is a simple and commonly used data augmentation method, which constructs a new samples by linear interpolation, and FedMix\cite{yoon2021fedmix} is another work that using Mixup strategies. However, Mixup cannot generate unseen labels. Unlike data augmentation, data sharing methods \cite{9412599, yoshida2020hybrid} alleviates the non-IID problem by collecting a small subset of samples from participants on the central server. However, these methods violate the privacy assumption in federated learning. Unlike these methods, our SL-GAN trains a global GAN for data augmentation.



\subsection{Federated Generative Models}


To synthesize fake data with a distribution similar to the global data, federated GAN \cite{rasouli2020fedgan, augenstein2019generative} train a global generative model among participants in federated learning. Existing federated GANs can be divided into two categories. One type is that generator trained on the server, discriminators are trained on the client. DP-FedAvg-GAN \cite{augenstein2019generative} is the first work in such architecture to train a global GAN for data generation, which under the assumption of IID data. F2U \cite{yonetani2019decentralized} follows previous settings and assign different weights for local discriminators in the process of model aggregation on the non-IID data. Another type \cite{rasouli2020fedgan} is that both discriminator and generator are trained on the client and aggregated on the server. Unfortunately, these methods require a central server and cannot guarantee the convergence of the GAN on non-IID data. In contrast to these methods, our SL-GAN trains a global GAN without a central coordinator. And we theoretically prove the convergence of SL-GAN.



\section{Proposed Methods}
\label{sec:sl-gan}

In this section, we propose SL-GAN framework, describe its training algorithm and theoretically prove its convergence. Table \ref{tab:symbols} summarizes notations used in this paper.

\begin{table}[tp]

  \centering
  \begin{threeparttable}
  \caption{Summary of notations}
  \label{tab:symbols}

    \begin{tabular}{cl}
    \toprule
    Notation & Description\cr
    \midrule
    $C$ & set of participants \cr
    $X$ & local dataset of participants \cr
    $lr$ & learning rate \cr
    $N$ & local training epochs \cr
    $T$ & aggregation time interval \cr
    $g_D, g_G$ & stochastic gradient of discriminator and generator \cr
    $\hat{g_D}, \hat{g_G}$ & true gradient of discriminator and generator \cr
    $\theta_D, \theta_G$ & parameters of discriminator and generator \cr
    $p$ & aggregation weight \cr
    $M^{\theta_D}, M^{\theta_G}$ & stochastic gradient errors\cr
    $v$ & discriminator parameters in the interval \cr
    $\phi$ & generator parameters in the interval \cr
    $\alpha$ & maximum round between two aggregations \cr
    $sup$ & supremum \cr
    $\mathbb{E}$ & expectation \cr
    $\mu$ & mean value \cr
    $\sigma$ & standard deviation \cr
     
    \bottomrule
    \end{tabular}
    \end{threeparttable}
\end{table}

\begin{figure*}[htb] 
\centering  
\includegraphics[width=0.9\linewidth]{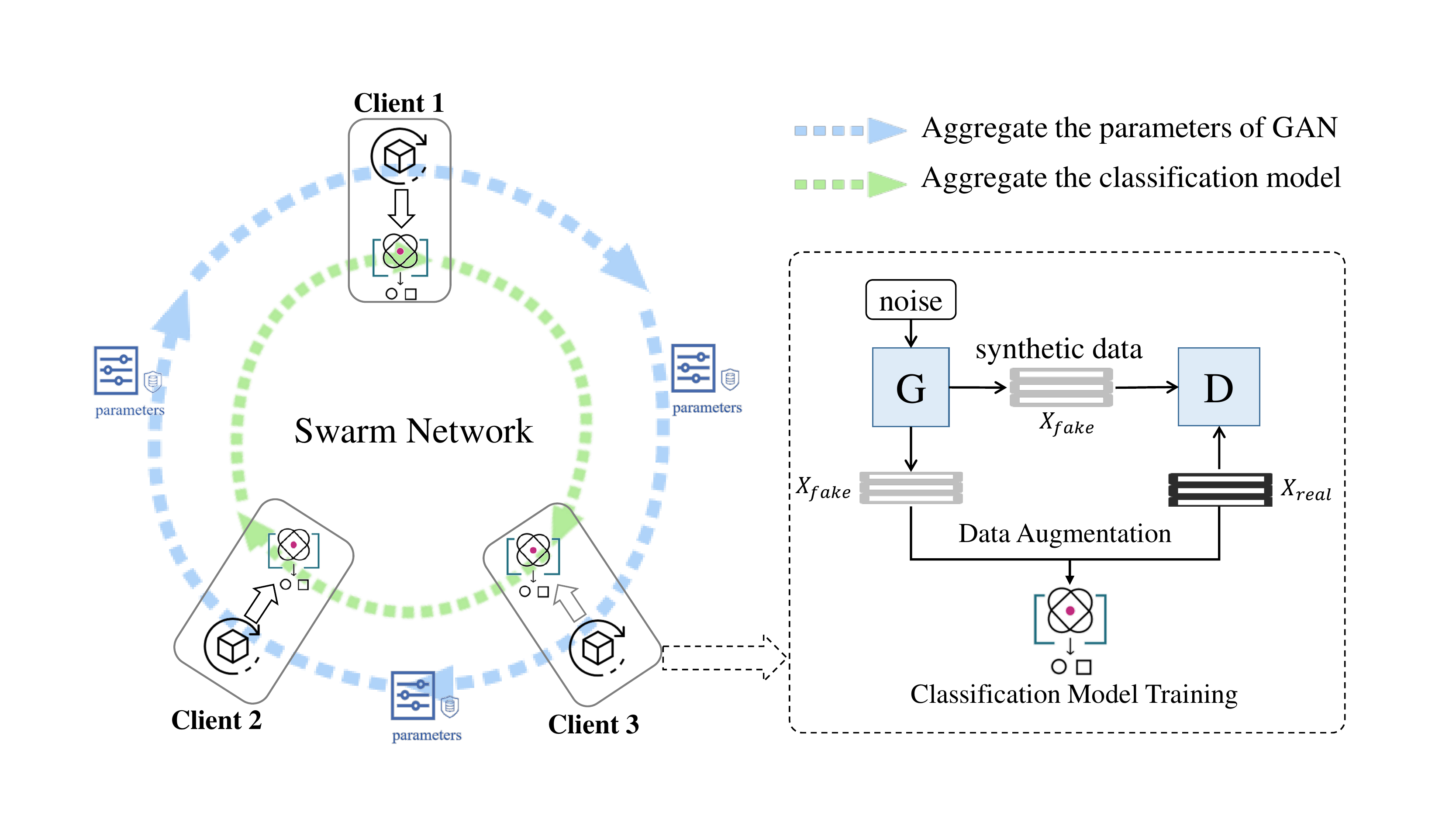} 
\caption{SL-GAN architecture. The green dotted arrow circle represents a swarm network used for aggregating the target clinical machine learning model. The blue dotted arrow circle denotes the swarm network that used to aggregate the parameters of GAN. Discriminators and generators are trained locally and aggregated on a elected edge node. Synthetic data generated by the trained generator are used to balance the data distribution among participants.}
\label{fig:architecture} 
\end{figure*}

\subsection{SL-GAN}

Figure \ref{fig:architecture} shows the architecture of our SL-GAN, in which jointly trains a GAN among participants. In SL-GAN, each participant trains generator and discriminator locally. When a user-defined
synchronization interval reached, the trained local discriminators and generators are aggregated on a randomly elected participant and then sent back to each participant with the swarm network. After SL-GAN converges, the trained generator is used for data augmentation. Finally, The target classification model is trained using the combination of the synthetic data and the local private data.




Algorithm \ref{alg:slgan} describes the training process of SL-GAN. 

\begin{enumerate}
\item[(1)] Each participant trains local generator and discriminator using standard GAN training procedure. The discriminator is trained using the real data $X_{real}$ and the fake data $X_{fake}$ generated by the generator, the generator is updated follow the discriminator (lines 1-6 of Algorithm \ref{alg:slgan}).
\item[(2)] When reached the pre-defined synchronization interval $T$, participants send their local discriminator and generator to a temporarily elected node for aggregation with the weight $p_{c}$, $p_{c} = \frac{|\mathcal{X}_c|}{\sum_{j \in C}|\mathcal{X}_j|}$ (lines 7-10 of Algorithm \ref{alg:slgan}).
\item[(3)] After model aggregation, each participant receives the aggregated discriminator and generator and updates their local model (line 11 of Algorithm \ref{alg:slgan}).
\end{enumerate}

The algorithm repeats the above process until SL-GAN converges.


    \begin{algorithm}[!t]
        \caption{SL-GAN model training process} 
        \label{alg:slgan}
        
        \begin{algorithmic}[1] 
            \Require 
            Local training epoch $N$, batch size $B$, learning rate of discriminator $lr_D(n)$, learning rate of generator $lr_G(n)$, local discriminator $\theta_{D_c}$, local generator $\theta_{G_c}$, synchronization interval $T$, start time $t_0$, current time $t$, weight of model aggregation $p_c$.
            \Ensure 
            well trained discriminator $\theta_{D_c}$ and generator $\theta_{G_c}$.
            
            \For{$n$ from $0$ to $N-1$ for all clients}
                \State $X_{real}^i \leftarrow$ (sample random batch data of batch size $B$) 
                \State $X_{noise}^i \leftarrow$ (sample random noise of batch size $B$) 
                \State $X_{fake}^i \leftarrow Generator(X_{noise}^i,\theta_{G}^i)$
                \State $\theta_{D}^i \leftarrow   \theta_{D}^i - lr_{D}(n)\bigtriangledown_{\theta_{D}^i}loss_{D}(\theta_{D}^i,X_{fake}^i,X_{real}^i)$
                \State $\theta_{G}^i \leftarrow   \theta_{G}^i - lr_{G}(n)\bigtriangledown_{\theta_{G}^i}loss_{G}(\theta_{G}^i,X_{fake}^i,\theta_{D}^i)$
                
                \If{$(t-t_0) \vert  T$}
                    \State Random select a participant $c'$ for model aggregation
                    \State $\theta_G^t \leftarrow \sum\limits_{c \in C}p_c\theta_{G_c}^t$ 
                    \State$\theta_D^t \leftarrow\sum\limits_{c \in C}p_c\theta_{D_c}^t$
                    \State Send back  $\theta_G^t, \theta_D^t$ and all participants update local discriminators and generators
                \EndIf
            \EndFor
            
        \end{algorithmic}
    \end{algorithm}
    
\subsection{Convergence Analysis}
In this section, we show that our SL-GAN converges in swarm learning with non-IID data. We denote the gradient of discriminator and generator in participant $i$ by $g_D^i$ and $g_G^i$. Let  $\theta^i = (\theta_D^i, \theta_G^i)^\top$ be the parameter of the participant $i$. The true gradient rather than stochastic gradient of each client is specified as $\hat{g_D^i}$ and $\hat{g_G^i}$. In addition, we define the stochastic gradient errors $M^{(g_D)}$ and $M^{(g_G)}$, where $M^{(\theta_D)} = \hat{g_D} - \Sigma_i p_i g_D^i$ and $M^{(\theta_G)} = \hat{g_G} - \Sigma_i p_i g_G^i$.

We follow the assumptions in the centralized GAN.
\begin{enumerate}
    \item $g_D^i$ and $g_G^i$ are $L$-Lipschitz.
    \item $\Sigma_n lr = \infty$, $\Sigma_n lr^2 < \infty$
    \item $\{M_n^{(\theta_D)}\}$ and $\{M_n^{(\theta_G)}\}$ are martingale difference sequence of the increasing $\sigma$-filed $\mathbb{F}_n=\sigma(\theta_{D_l}, \theta_{G_l}, M_l^{(\theta_D)}, M_l^{(\theta_G)}, l \leq n), n \geq 0$.
    \item $sup_{n} ||\theta_{D_{n}}|| \textless \infty $ and $sup_{n} ||\theta_{G_n}|| \textless \infty $
    \item $\mathbb{E}||g_D^i-\hat{g_D^i}||\leq \sigma_{g_D}, \mathbb{E}||g_G^i-\hat{g_G^i}||\leq \sigma_{g_G}$ and $||g_D^i - \hat{g_D^i}|| \leq \mu_{g_D}$
\end{enumerate}
where (1)-(4) are used in stochastic approximation of GAN convergence. In assumption (5), the first bound ensures that the local stochastic gradient is close to the local true gradient, the second bound ensures that the local discriminator true gradient of the non-IID data are close to the discriminator true gradient of the pooled data \cite{rasouli2020fedgan}, the last bound represents bounded gradient divergence.

To prove the convergence of SL-GAN, we connects the convergence of GAN to the convergence of an ordinary differential equations (ODE) representation of the parameter updates \cite{mescheder2017numerics}. We prove that the ODE representing the parameter updates of SL-GAN asymptotically tracks the ODE representing the parameter updates of the centralized GAN. As defined in \cite{mescheder2017numerics, nagarajan2017gradient}, the centralized GAN tracks the following ODE asymptotically.


\begin{equation} \label{ODE}
\dot{\theta}=
\left(\begin{array}{c}
\dot{\boldsymbol{\theta_D}}(t) \\
\dot{\boldsymbol{\theta_G}}(t)
\end{array}\right)=\left(\begin{array}{l}
\boldsymbol{g_D}(\boldsymbol{\theta_D}(t), \boldsymbol{\theta_G}(t)) \\
\boldsymbol{g_G}(\boldsymbol{\theta_D}(t), \boldsymbol{\theta_G}(t))
\end{array}\right) \cdot
\end{equation}

Therefore, the problem of proving the convergence of SL-GAN is transformed into proving that the parameter of SL-GAN follows (\ref{ODE}) asymptotically. As proofed in \cite{rasouli2020fedgan}, we have,


\begin{flalign}
\label{equ:l1}
& \mathbb{E}\left\|\boldsymbol{\theta_D}_{n}^{i}-\boldsymbol{v}_{n}\right\|+\mathbb{E}\left\|\boldsymbol{\theta}_{G}^{i}-\boldsymbol{\phi}_{n}\right\| \leq  \nonumber \\
& \frac{\sigma_{\theta_D}+\mu_{\theta_D}+\sigma_{\theta_G}}{2 L}\left[(1+2 lr(n-1) L)^{n \bmod K}-1\right] 
\end{flalign}

\begin{flalign}
\label{equ:l2}
& \mathbb{E}\left\|\boldsymbol{\theta_D}_{n}-\boldsymbol{v}_{n}\right\|+\mathbb{E}\left\|\boldsymbol{\theta}_{G}-\boldsymbol{\phi}_{n}\right\| \leq \nonumber \\
&\frac{\left(\sigma_{\theta_D}+\mu_{\theta_D}+\sigma_{\theta_G}\right)}{2 L}\left[(1+2 lr(n-1) L)^{K}-1\right]-lr(n-1) \mu_{\theta_D} K 
\end{flalign}

$v_n$ and $\phi_n$ represent the parameter of discriminator and generator in the interval between two aggregations, respectively. The specific definition is as follows
\begin{align}
    \boldsymbol{v}_{n}=\boldsymbol{\theta_D}_{n_{1}}+\sum_{k=n_{1}}^{n} lr(k) \boldsymbol{g_D}\left(\boldsymbol{\phi}_{k}, \boldsymbol{v}_{k}\right),  \\ \boldsymbol{\phi}_{n}=\boldsymbol{\theta_G}_{n_{1}}+\sum_{k=n_{1}}^{n} lr(k) \boldsymbol{g_G}\left(\boldsymbol{\phi}_{k}, \boldsymbol{v}_{k}\right) 
\end{align}
where $n_1$ means the nearest aggregation timestamp. However, in swarm learning, there is no $K$, which indicates every $K$ local epochs. Fortunately, we can still prove that in the interval between every two adjacent aggregations, the local epochs for every client is bounded, 
$$n^i - n_1 \leq \alpha, \forall i \in \{1, \cdots , m\} \quad \alpha > 0$$
which is trivial because a participant will not train itself infinitely anyway. We just need to treat $max(n^i - n_1)$ as $K$, bringing it to the Equations (\ref{equ:l1}) and (\ref{equ:l2}), we will get the same result as federated learning. Based on the Theorem 1 in \cite{rasouli2020fedgan} and Theorem 2 in  \cite{borkar2009stochastic}, $\theta$ in SL-GAN tracks the ODE (\ref{ODE}) asymptotically, namely it will converge eventually.

\section{Experiments}
\label{sec:exp}


\begin{table*}[h]
    \centering
    \renewcommand\arraystretch{1.5}
    \caption{Statistics of Datasets}
    \label{tab:dataset}
    \begin{tabular}{cccccc}
        \toprule 
        Dataset & \makecell[c]{$\sharp$\ of samples\\(training set)} & \makecell[c]{$\sharp$\ of samples\\(test set)} & $\sharp$\ of columns & \makecell[c]{Distribution of samples\\(training set)\\(CASE:CONTROL)} & \makecell[c]{Distribution of samples\\(test set)\\(CASE:CONTROL)} \\
        \midrule
        \makecell[c]{Tuberculosis} & 1240 & 310 & 18136 & 620:620 & 155:155 \\
        \makecell[c]{Leukemia} & 1943 & 436 & 22283 & 826:1117 & 206:230 \\
        \makecell[c]{COVID-19} & 1920 & 480 & 19400 & 237:1683 & 59:421 \\
        \bottomrule
    \end{tabular}
\end{table*}

\subsection{Experimental Setup}
\subsubsection{Datasets}

We use three real-world clinical datasets, as shown in Table \ref{tab:dataset}, and their details are as follows: 
\begin{enumerate}
\item[$\bullet$] Tuberculosis \cite{warnat2021swarm} is an RNA-Seq dataset based on whole blood transcriptomes, which combines data from healthy controls with published data in Gene Expression Omnibus (GEO). This dataset merged from 9 independent datasets: GSE101705, GSE107104, GSE112087, GSE128078, GSE66573, GSE79362, GSE84076, GSE89403. There are 1550 samples from patients with active tuberculosis, latent tuberculosis, fatigue, autoimmune diseases, HIV and controls. All active tuberculosis samples are listed as CASE and and all other samples are listed as CONTROL. After data preprocessing, there are 18136 genes (columns) in this dataset. 


\item[$\bullet$] Leukemia dataset \cite{warnat2020scalable} contains 2379 transcriptomes derived from peripheral blood mononuclear cells (PBMC) or bone marrow, published at the GEO under subseries GSE122505. In this dataset, independent data sets selected from GEO are as follows: GSE10255, GSE1159, GSE12417, GSE12995, GSE13425, GSE14471, GSE14895, GSE16129, GSE25571, GSE26281, GSE33315, GSE34860, GSE37642, GSE43176, GSE4698, GSE51082, GSE6269, GSE67684, GSE83449, GSE8879, GSE9006, GSE9476. Acute Myeloid Leukemia (AML) samples are classified as CASE and all other samples are classified as CONTROL.

\item[$\bullet$] COVID-19 \cite{warnat2021swarm} is an RNA-Seq dataset based on whole blood transcriptomes, which contains 296 samples from patients with COVID-19, as well as 2104 other control samples (autoimmune disease, Fatigue, healthy controls, HIV, latent tuberculosis and active tuberculosis). COVID-19 samples are labeled as CASE and all other samples are labeled as CONTROL. After data preprocessing, there are 19400 genes (columns) in this dataset. 

\end{enumerate}

Table \ref{tab:dataset} shows the statistical information of three datasets. We split the total samples into
training dataset and test dataset with the ratio 8:2.

\subsubsection{Baseline Methods} We adapted two state-of-the-art  algorithm-based methods (Fedprox, FedNova) in federated learning and one data-based method (Mixup) to swarm learning as baseline methods. The detail information of baseline methods are as follows:

\begin{enumerate}


\item[--] \textbf{Fedprox} \cite{li2020federated} is an algorithm-based method, which modify the objective function of participant $k$ as follows, 
\begin{equation} 
h_{k}\left(w ; w^{t}\right)=F_{k}(w)+\frac{\mu}{2}\left\|w-w^{t}\right\|^{2}
\label{equ:fedprox}
\end{equation}
where $F_{k}(w)$ represents the original objective function, $w^t$ is the parameter of the global model obtained in the $t$-th round. 

\item[--] \textbf{FedNova} \cite{wang2020tackling} is an algorithm-based method, which modify the original aggregation method to
\begin{equation} 
w^{t+1}-w^{t}=(\dfrac{\sum_{k=1}^K p_k\tau_k^{(t)}}{\tau_k^{(t)}}) \sum_{k=1}^K p_k\Delta_k^{(t)}
\label{equ:fednova}
\end{equation}
where $\tau_k^{(t)}$ is the number of iterations of participant $k$ in round $t$, and $\Delta_k^{(t)}$ is the gradient of participant $k$ in round $t$.

\item[--] \textbf{Mixup} \cite{zhang2018mixup}
is a simple data augmentation method,
\begin{equation} 
\left\{\begin{array}{l}
\tilde{x}=\lambda x_{i}+(1-\lambda) x_{j} \\
\tilde{y}=\lambda y_{i}+(1-\lambda) y_{j}
\end{array}\right.
\label{equ:mixup}
\end{equation}
where $(x_i,y_i)$ and $(x_j,y_j)$ are two randomly selected samples, $\lambda\in(0,1)$.
\end{enumerate}

\subsubsection{Data Partition}
To simulate the non-IID data in real world, we use the distributed-based label imbalance method in \cite{li2021federated}, that each participant is allocated a proportion of the samples of each label according to Dirichlet distribution. In practice, we allocate a $p_{k,j}$ ($p_{k,j} \sim Dir(\beta)$) proportion of the samples of label $k$ to participant $j$, where $Dir(.)$ represents Dirichlet distribution and $\beta$ ($\beta > 0$) is its parameter. In this approach, we can flexibly control the degree of the non-IID by varying the parameter $\beta$. The smaller the $\beta$ is, the more unbalanced the data distribution is.


\subsubsection{Data Augmentation}
 After SL-GAN converges, the synthetic data generated by the trained generator is used to augment local data based on global data distribution. During training, each participant contains the same amount of data and its  label distribution is consistent with the global label distribution. For instance, the data contained in a participant is \{CONTROL:40, CASE:414\} on the Tuberculosis dataset with $\beta=1$, and the distribution of global data is CONTROL:CASE = 1:1. After data augmentation, the data of the three participants are \{CONTROL:556, CASE:556\}.

\subsection{Performance on Classification}

\begin{figure*}[!t] 
\centering 
\includegraphics[width=0.7\linewidth]{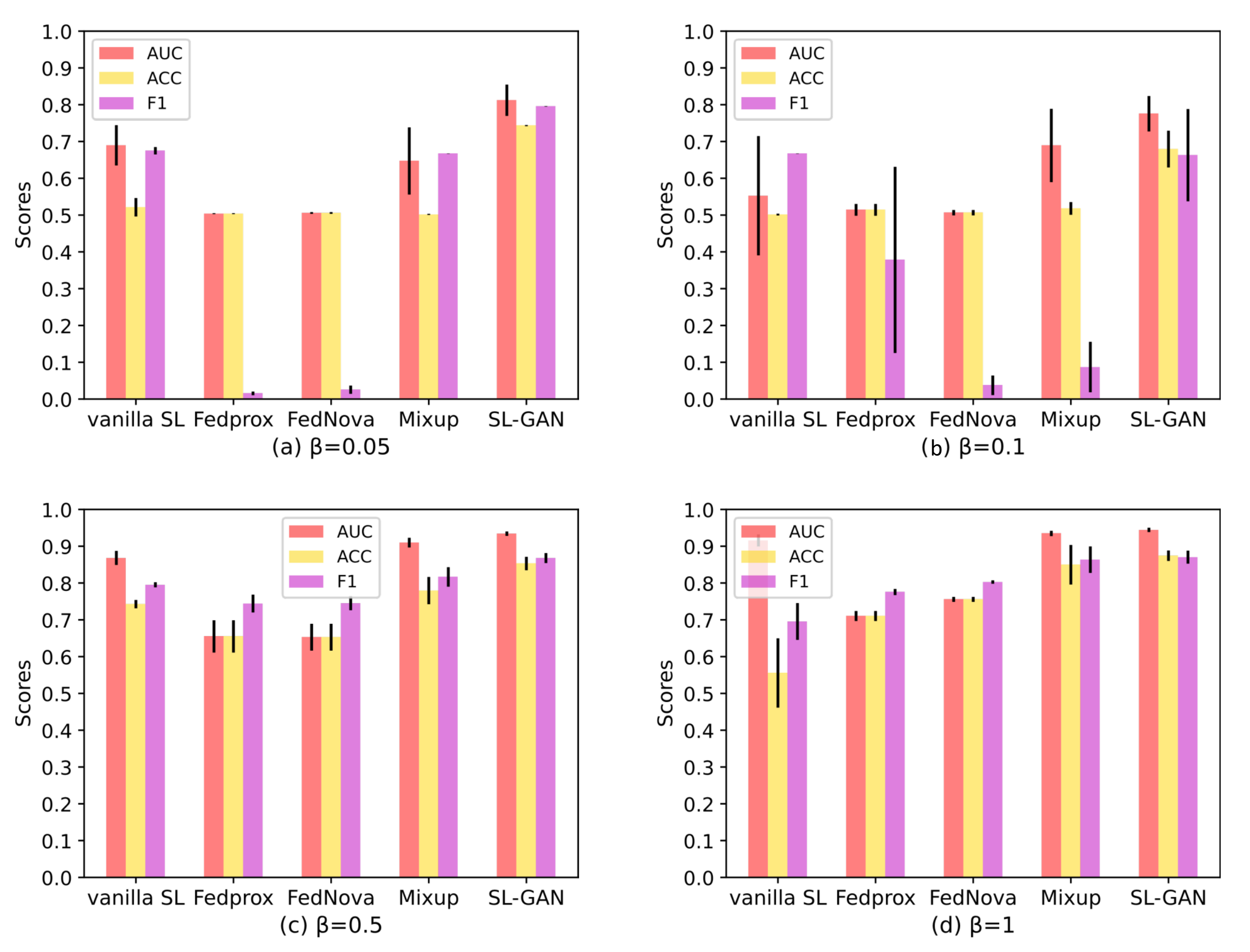} 
\caption{Performance on the Tuberculosis dataset. We compared with vanilla swarm learning, Fedprox, FedNova, Mixup and SL-GAN in terms of AUC, F1 and accuracy scores. The red bar represents AUC scores, the yellow bar denotes accuracy, and the purple bar represents F1 scores.}
\label{fig:tuberculosis} 
\end{figure*}

\begin{figure*}[!t] 
\centering 
\includegraphics[width=0.7\linewidth]{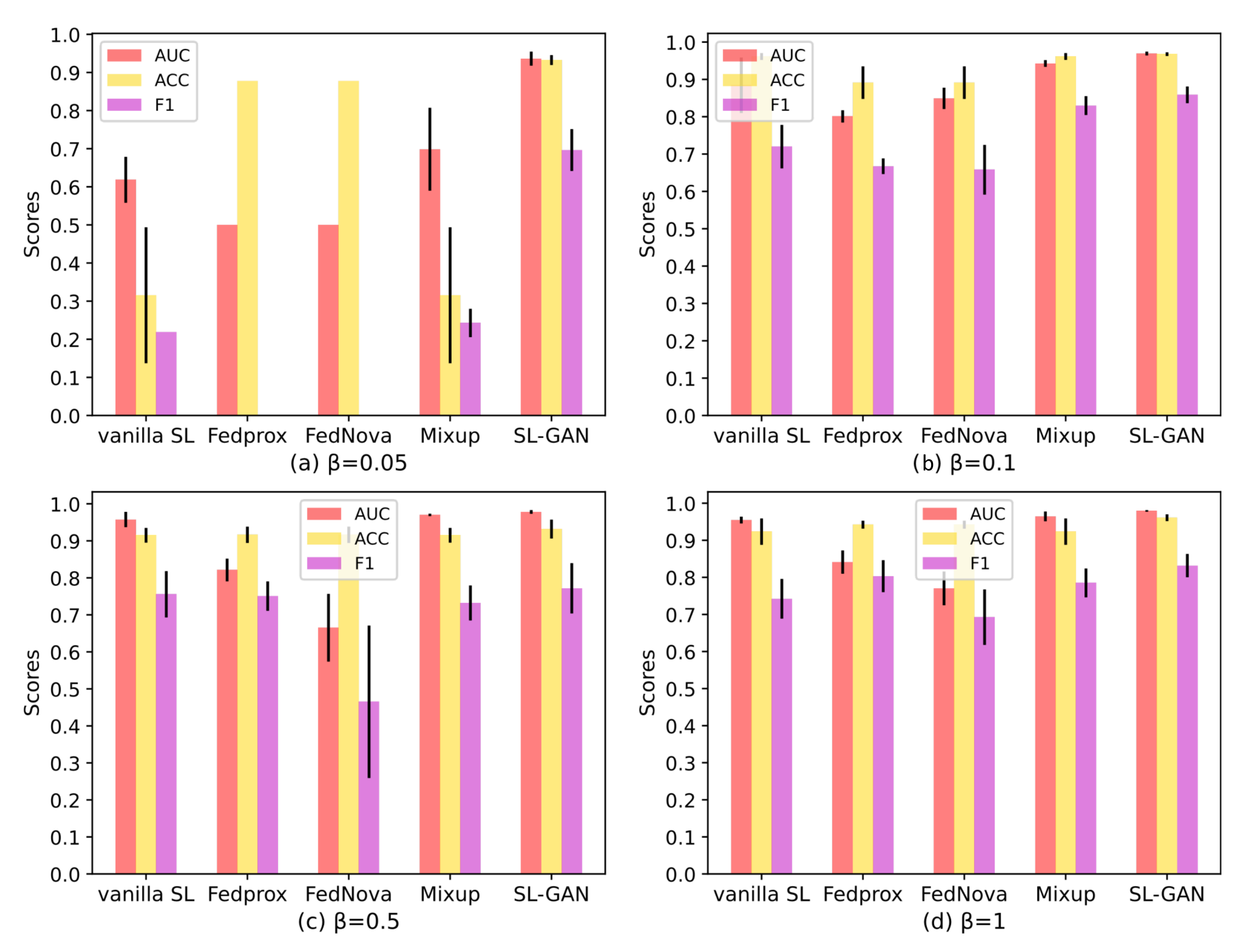} 
\caption{Performance on the COVID-19 dataset. We compared with vanilla swarm learning, Fedprox, FedNova, Mixup and SL-GAN in terms of AUC, F1 and accuracy scores. The red bar represents AUC scores, the yellow bar denotes accuracy, and the purple bar represents F1 scores.}
\label{fig:covid19} 
\end{figure*}

\begin{figure*}[!t] 
\centering 
\includegraphics[width=0.7\linewidth]{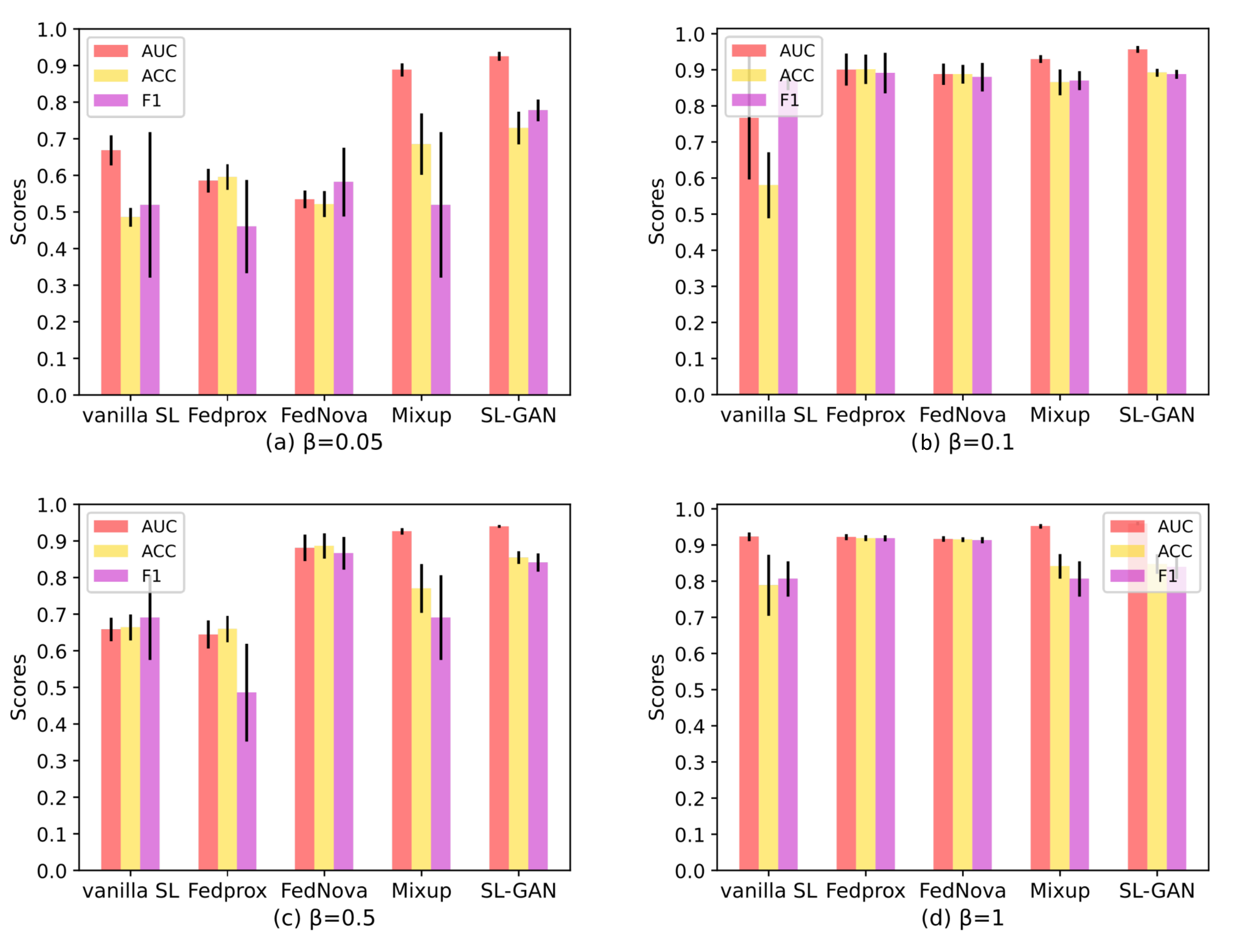} 
\caption{Performance on the Leukemia dataset. We compared with vanilla swarm learning, Fedprox, FedNova, Mixup and SL-GAN in terms of AUC, F1 and accuracy scores. The red bar represents AUC score, the yellow bar denotes accuracy, and the purple bar represents F1 score.}
\label{fig:leukemia} 
\end{figure*}

\begin{figure*}[!t] 
\centering 
\includegraphics[width=0.7\linewidth]{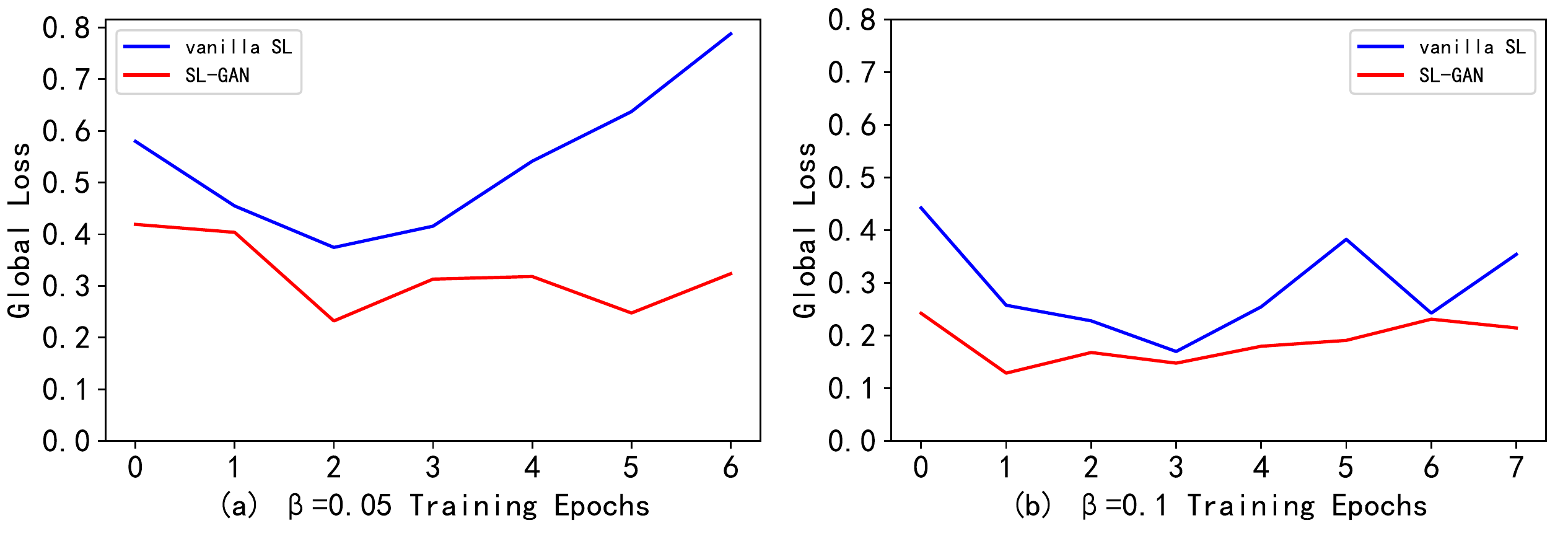} 
\caption{Communication round on the COVID-19 dataset. The blue lines represent the training round of the vanilla SL algorithm, and the red lines represent the training round of SL-GAN}
\label{fig:loss} 
\end{figure*}

Figures \ref{fig:tuberculosis} - \ref{fig:leukemia} show the comparisons of the performance on Tuberculosis, COVID-19, Leukemia dataset, respectively. We compare vanilla swarm learning, Fedprox, FedNova, Mixup with our SL-GAN in four different data distributions ($\beta = 0.05, 0.1, 0.5, 1$). SL-GAN  outperforms all the baseline methods in terms of F1 score, accuracy, and AUC in three datasets. 

In many cases, the vanilla SL algorithm outperforms Fedprox and FedNova, which is consistent with the conclusion in \cite{li2021federated}. This is because algorithm-based methods do not fundamentally address the problem of data imbalance. In contrast, Mixup and SL-GAN both perform better than the vanilla SL algorithm in almost all cases, which means that data augmentation methods have a significant effect on alleviating data imbalance. Unfortunately, in Figure \ref{fig:tuberculosis} (a) and Figure \ref{fig:covid19} (a), F1 scores of Fedprox and FedNova are close to 0, which suggests that these methods fail to make the local model consistent with the global model under extreme data imbalances. Compared with Mixup, SL-GAN shows more reliable performance in various distributions. This is because Mixup cannot generate the class of samples that are not available in the local. Therefore, in some cases, the enhancement is redundant data features. SL-GAN trains a generative model among participants and learns the real distribution, which can generate synthetic data with an approximate distribution of the global data. To summarize, in almost all cases, SL-GAN achieves the best performance. In general, data-based methods show better performance than algorithm-based method because they construct a balanced data distribution among each participant.

Figure \ref{fig:loss} shows the training round on the COVID-19 dataset. As we can see, by augmenting the synthetic data, SL-GAN converges to a smaller loss than the vanilla SL algorithm. As shown in Figure \ref{fig:loss}, with the degree of non-IID increases ($\beta$ form 0.1 to 0.05), the vanilla SL algorithm has difficulty in convergence. In contrast, the model augmented by SL-GAN can still converge to a relatively small loss, which means that the data augmentation method proposed in this paper can significantly improve the efficiency and effect of the model training.

\subsection{Synthetic Data Utility}

\begin{table}[!t]
\label{tab:synthetic}
    \centering
    \renewcommand\arraystretch{1.5}
    \caption{The accuracy of the synthetic data on the COVID-19 dataset}
    \resizebox{\linewidth}{!}{
    \begin{tabular}{ccccccc}
        \toprule 
        Classifier& Original data & $\beta$\ = 1 & $\beta$\ = 0.5 & $\beta$\ = 0.1 & \makecell[c]{$\beta$\ = 0.05} \\
        \midrule
        \makecell[c]{LGBMClassifier} & 0.98 & 0.74 & 0.72 & 0.83 &	0.58\\
        \makecell[c]{XGBClassifier} & 0.97	& 0.70 & 0.69 &	0.72 & 0.60 \\
        \makecell[c]{BaggingClassifier} & 0.97	& 0.84	& 0.76 &	0.84 &	0.79 \\
        \makecell[c]{SVC} & 0.96 &	0.88 &	0.88 &	0.88 &	0.88 \\
        \makecell[c]{RandomForestClassifier} & 0.96 &	0.85 & 0.86 &	0.86 &	0.82 \\
        \makecell[c]{LabelPropagation} & 0.88 & 0.88 &	0.88 &	0.88 &	0.88 \\
        \makecell[c]{ExtraTreesClassifier} & 0.96 &	0.84 &	0.84 &	0.84 &	0.85 \\
        \makecell[c]{CalibratedClassifierCV} & 0.96 &	0.84 &	0.88 &	0.88 &	0.87 \\
        \makecell[c]{GaussianNB} & 0.83 &	0.86 &	0.88 &	0.80 &	0.86 \\
        \makecell[c]{LabelSpreading} & 0.88 & 0.88 &	0.88 &	0.88 &	0.88 \\
        \makecell[c]{LabelPropagation} & 0.88 & 0.88 &	0.88 &	0.88 &	0.88 \\
        \hline
        \makecell[c]{Average} & 0.935 &	0.831 &	0.827 &	0.841 &	0.801 \\
        \bottomrule
    \end{tabular}}
\end{table}

To evaluate the utility of the synthetic data that generated by SL-GAN, we train 10 machine learning models on the synthetic data and test these models on the real data. The performance of the synthetic data for COVID-19 dataset shown in Table \ref{tab:synthetic}. As shown in Table \ref{tab:synthetic}, as the value of $\beta$ decreases (from 1 to 0.05), the accuracy of the synthetic data does not change. This shows that the proposed SL-GAN can converge stably on the non-IID data. In many cases, the performance of the synthetic data is close to that of the original data. However, The average accuracy of the SL-GAN is 10\% lower than that of the original data, which means that our generative method does not always successfully capture the real features of the original data.

\section{Conclusion and Future Works}
\label{sec:conclusion}

In this paper, we presented a generative augmentation framework in swarm learning for non-IID data, which called SL-GAN. We jointly train a GAN in the swarm learning network, and theoretically prove the convergence of the SL-GAN. We augments the non-IID data into IID data using SL-GAN. We evaluated the proposed SL-GAN on three real-world clinical dataset. The experimental results show that our SL-GAN outperforms the state-of-the-art work in various data distributions.

In the future, in order to protect the data privacy, the differential privacy will be introduced to the SL-GAN and the privacy of the synthetic data will be studied. Furthermore, as there is still a gap between the utility of the synthetic data and original data, we will combines prior knowledge to improve the quality of the synthetic data.

\section*{Acknowledgment}

This work was supported by the National Natural Science Foundation of China Grant No.61872110.


\bibliographystyle{IEEEtran}
\bibliography{IEEEabrv,refs}






\end{document}